\documentclass{article}

\usepackage{arxiv}
\usepackage{iccv}

\usepackage[utf8]{inputenc} % allow utf-8 input
\usepackage[T1]{fontenc}    % use 8-bit T1 fonts
\usepackage{hyperref}       % hyperlinks
\usepackage{url}            % simple URL typesetting
\usepackage{booktabs}       % professional-quality tables
\usepackage{amsfonts}       % blackboard math symbols
\usepackage{nicefrac}       % compact symbols for 1/2, etc.
\usepackage{microtype}      % microtypography
\usepackage{times}
\usepackage{epsfig}
\usepackage{graphicx}
\usepackage{amsmath}
\usepackage{amssymb}

\newcommand{\repeatthanks}{\textsuperscript{\thefootnote}}

\DeclareUnicodeCharacter{3000}{}

\title{Hard-Aware Fashion Attribute Classification}

\author{
  Yun Ye\thanks{This work was done while Yun Ye \& Yixin Li was with JD AI Research.} \\
  Intel\\
  \texttt{yun.ye@intel.com} 
   \And
 Yixin Li\repeatthanks \\
  Peking University \\
  \texttt{liyixin@pku.edu.cn} 
  \And
 Bo Wu \\
  Columbia University \\
  \texttt{bo.wu@Columbia.edu} 
  \And
 Wei Zhang \\
  JD AI Research \\
  \texttt{zhangwei96@jd.com} 
    \And
 Lingyu Duan \\
  Peking University \\
  \texttt{lingyu@pku.edu.cn} 
    \And
 Tao Mei \\
  JD AI Research \\
  \texttt{tmei@jd.com} 
}

\begin{document}
\maketitle

\begin{abstract}
Fashion attribute classification is of great importance to many high-level tasks such as fashion item search, fashion trend analysis, fashion recommendation, etc. The task is challenging due to the extremely imbalanced data distribution, particularly the attributes with only a few positive samples. In this paper, we introduce a hard-aware pipeline to make full use of ``hard'' samples/attributes. We first propose Hard-Aware BackPropagation (HABP) to efficiently and adaptively focus on training ``hard'' data. Then for the identified hard labels, we propose to synthesize more complementary samples for training. To stabilize training, we extend semi-supervised GAN by directly deactivating outputs for synthetic complementary samples (Deact). In general, our method is more effective in addressing ``hard" cases. HABP weights more on ``hard" samples. For "hard" attributes with insufficient training data, Deact brings more stable synthetic samples for training and further improve the performance. Our method is verified on large scale fashion dataset, outperforming other state-of-the-art without any additional supervisions.
\end{abstract}

%%%%%%%%% BODY TEXT
\section{Introduction}

\begin{figure}[ht]
	\begin{center}
		\includegraphics[width=0.8\linewidth]{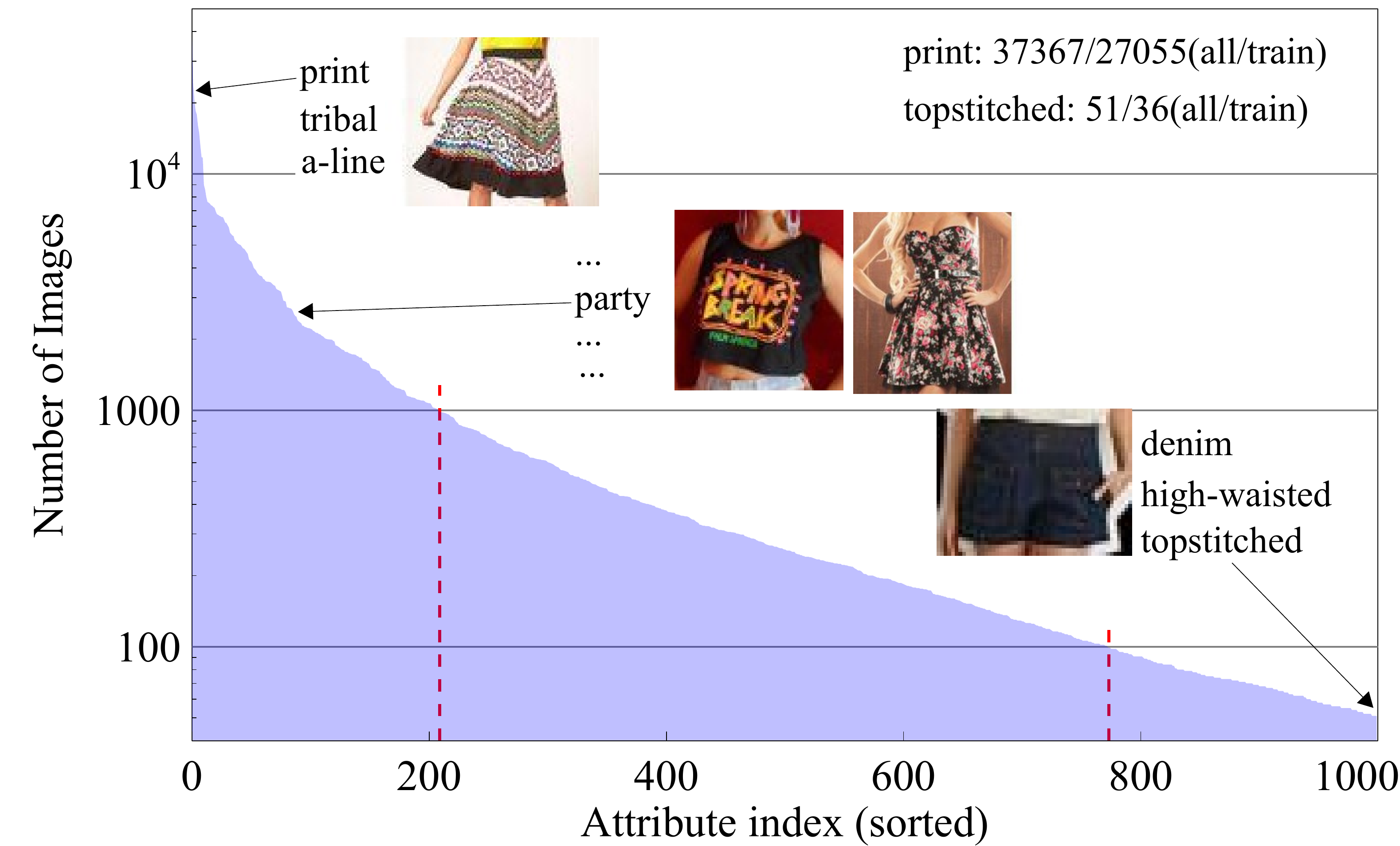}
	\end{center}
	\caption{Sample images and statistics of DeepFashion-C. Over 1/5 attributes have fewer than 100 positive samples. $\sim$80\% attributes has fewer than 1000 positive samples. }
	\label{fig:DFC}
\end{figure}

Attributes, also known as mid-level semantic features~\cite{5206772,6248089}, is fundamental for describing fashion items. As an example, in Fig. \ref{fig:DFC}, the skirt shown in the upper plot can be described with ``print" texture, ``tribal" style and ``a-line" shape. Attributes have been extensively used in many computer vision tasks, such as image retrieval~\cite{5995329,6419834}, person Re-ID~\cite{10.1007/978-3-642-33863-2_40}, etc. Particularly in fashion domain, cloth attribute is of great importance to many other high-level tasks including fashion image classification~\cite{7780493, 7298688}, fashion item search~\cite{7410739,7410484,7780493,7299169,6248071}, fashion style understanding~\cite{Ma:2017:TBU:3298239.3298246,DBLP:journals/corr/MatzenBS17,8265473,8237713}, fashion recommendation~\cite{7298688,Han:2017:LFC:3123266.3123394,Liu:2012:HMC:2393347.2393433}, fashion outfit learning~\cite{Han:2017:LFC:3123266.3123394,Song:2017:NNC:3123266.3123314, Hsiao_2018_CVPR}, and fashion trend analysis~\cite{Chen:2015:DWP:2733373.2809930,7298688,Al-Halah_2017_ICCV}. 

\begin{figure}[ht]
	\begin{center}
		\includegraphics[width=0.8\linewidth]{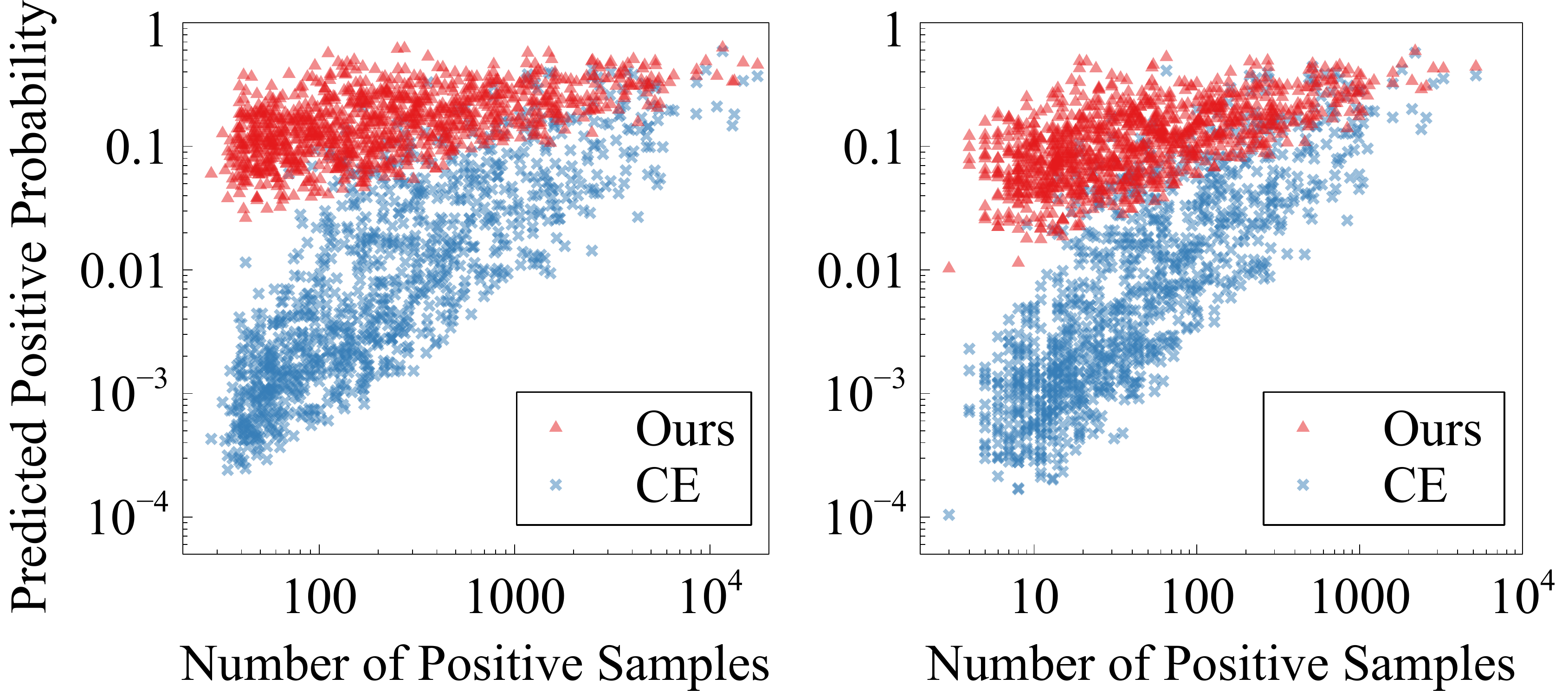}
	\end{center}
	\caption{Model predicted probabilites for positive samples vs. numbers of positive samples on DeepFashion-C. CE: Cross Entropy loss. {\bf Left}: Train set; {\bf Right}: Test set. In general, our method handles better on attributes with only a few positive samples. }
	\label{fig:ppcorr}
\end{figure}

In this paper, we address one of the major problems in fashion attribute classification: imbalanced data distribution, specifically the samples or attributes with very few positive labels. Patterns in fashion images are highly diversified due to its non-rigid nature and abundant semantic behind. Combined with very rich attributes of fashion items, it brings the imbalance and sparsity of positive labels for some attributes or specific kind of samples. The upper plot in Fig. \ref{fig:DFC} demonstrates the positive attribute counts from DeepFashion: Category and Attribute Prediction Benchmark (DeepFashion-C)~\cite{7780493}. The dataset contains images and tags from shopping websites and search engine, that is representative in a real-world scenario. Among the 1000 annotated attributes, the most frequent label ``print'' has 37,367 occurrences, whereas the least label ``topstitched" only shows up in 51 images. In addition to imbalance, fashion attributes are usually sparsely distributed as shown in Fig. \ref{fig:DFC}, over 1/5 attributes have fewer than 100 positive labels and on average there are only 3.3 positive tags per image. Moreover, the diversity of fashion items makes the problem even worse. Take ``party" as an example, countless diversified fashion images can be defined as ``party" (Fig. \ref{fig:DFC}), such that a specific minority ``party" case may not be easy to learn from the 2,882 tagged samples. So the problem is at both attribute and sample level. A big difficulty in training with such kind of dataset is that majority data are generally well trained while minority data is either under-trained or is prone to over-fitting with too few samples. 

Many efforts have been devoted to tackling this problem~\cite{Krawczyk2016}. A common solution is re-sampling~\cite{Drummond2003C4,Chawla:2002:SSM:1622407.1622416,McCarthy:2005:CLB:1089827.1089836}. Though has been widely used, over-sampling has its limitations such as the tendency to over-fit, whereas under-sampling suffers from the risk of missing valuable information. Moreover, it is not trivial to extend re-sampling to multi-label datasets~\cite{Charte:2015:MLS:2839534.2840113,doi:10.1093/bioinformatics/btw560,DBLP:conf/aaai/HandCC18}, and few of them focused on imbalanced multi-label computer vision problems~\cite{8353718}. Another popular family takes into account the misclassification errors, known as cost-sensitive learning~\cite{Drummond2003C4,1549828,8012579,Kukar98cost-sensitivelearning, Ling2010}. Broadly speaking, it covers a wide range of methods that use algorithms or strategies based on cost. Among the scope, hard-aware methods are being actively studied in recent years with deep neural networks, such as focal loss~\cite{Lin_2017_ICCV}, hard example mining~\cite{Shrivastava_2016_CVPR,Yuan_2017_ICCV,8353718}, etc. 

In this work, we develop an approach leveraging both cost-sensitive and re-sampling strategies to make full use of ``hard'' data. The key idea is to focus on the minority data as much as possible, and don't affect the majority since they usually are already well trained. Minority data are often strongly correlated with high classification error as suggested by \cite{Chawla:2002:SSM:1622407.1622416,Lin_2017_ICCV}. We also verified this by comparing the average predicted probability for positive attributes vs. the number of positive labels in Fig. \ref{fig:ppcorr} (more details will be discussed in Section \ref{sec:maineval}). In the figure, the blue crosses are predicted probabilities for positive samples, from a well-trained model using cross entropy loss. Based on this, we use the error probability estimated by model~\cite{Lin_2017_ICCV} as a metric to identify ``hard'' data. To make the best of this key metric in training, two techniques are developed. We first present a solution from the view of cost-sensitive learning that to backpropagate losses on each sample and each attribute weighted by the estimated errors. We refer this method as Hard-Aware BackPropagation (HABP). From the perspective of re-sampling, we further suggest to sample synthetic complementary images, which are samples that around but not overlap with real samples in feature space, to train hard/minority attributes with generative adversarial networks~\cite{NIPS2014_5423} (GAN). The proposed method is similar to semi-supervised GAN~\cite{DBLP:journals/corr/Odena16a} but is much easier to train and implement. A possible reason that GAN is not widely used in a practical problem is the trickiness to train with high-resolution such as $224\times 224$. This was induced by problems including mode collapse~\cite{DBLP:journals/corr/RadfordMC15} and gradient vanishing~\cite{DBLP:journals/corr/ArjovskyB17}. In order to generate diversified high-resolution complementary images, we introduce a decorrelation regularization loss to deal with mode collapse. It successfully relieves mode collapse in training a multi-resolution GAN (MR-GAN) architecture we used in this work. 

Evaluations on DeepFashion-C demonstrates that our approach outperforms the state-of-the-art, without using additional supervisions. Our main contribution is proposing to take full advantage of ``hard'' samples with two techniques from the view of cost-sensitive learning and re-sampling respectively: 1) We propose Hard-Aware BackPropagation (HABP) that effectively reduce the impact of strong imbalance in multi-label image dataset. 2) Based on hard labels identified, we present a method to train model with synthetic complementary samples and a decorrelation loss for stably generating high-resolution synthetic samples.

%------------------------------------------------------------------------
\section{Related Work}

%-------------------------------------------------------------------------
\subsection{Fashion Attribute Classification}

Fashion attribute classification has already become a prevalent topic in the research area~\cite{7410739}. However in the early stage, most published datasets are either small-scale or annotated with a few numbers of attributes~\cite{10.1007/978-3-642-33712-3_44,10.1007/978-3-642-37447-0_25,7299169}. Based on DeepFashion-C, FashionNet~\cite{7780493} proposed to jointly learn cloth attributes and landmarks. Corbière \etal~\cite{Corbiere_2017_ICCV} collected noisy data from shopping website to perform weakly supervised image tagging. In the recent work \cite{Wang_2018_CVPR}, the authors grounded human knowledge to landmark detection. Then attribute classification was improved with landmark enhanced visual attention. Most existing works incorporated other supervision (such as landmarks, low-level features~\cite{10.1007/978-3-642-33712-3_44}) to improve attribute classification. A few of them~\cite{8353718} used attribute annotations only, but the method is not strongly tied to vision problems. In contrast, our method only uses attribute annotations and makes full application of training images in a semi-supervised manner. 

%-------------------------------------------------------------------------
\subsection{Hard-Aware Learning}

Hard example mining~\cite{Suykens1999} has been making successes with deep neural networks in areas including face recognition~\cite{Schroff_2015_CVPR}, object detection~\cite{Shrivastava_2016_CVPR}, person Re-ID~\cite{DBLP:journals/corr/HermansBL17}, and metric learning~\cite{Yuan_2017_ICCV}. Based on the same idea that hard samples are usually more informative, variants have been proposed. Among them, focal loss~\cite{Lin_2017_ICCV} (FL) is closely related to our work by sharing the idea of modeling the estimated probability of classification error and take it as weights in loss function. Variants of FL has been applied to attribute classification~\cite{Sarafianos_2018_ECCV}. A key difference between HABP and FL is that HABP introduces an output dependent normalization term for better stability and performance. OHEM~\cite{Shrivastava_2016_CVPR} is also related to our method in the idea of sampling ``hard'' data. More details will be discussed in Section \ref{sec:methodoloy}. 

\subsection{GAN \& Semi-Supervised GAN}

GAN~\cite{NIPS2014_5423,DBLP:journals/corr/RadfordMC15} has enjoyed a resurgence of interest in recent years for its ability to generate high fidelity images. A number of efforts have been made for synthesizing higher-resolution images. Denton \etal~\cite{NIPS2015_5773} employed a Laplacian pyramid with multiple discriminators to generate images at multiple resolutions. The idea of multi-resolution was further developed in \cite{karras2018progressive} with the progressive growth of GAN. Based on the idea of weight sharing across multiple resolutions, Karnewar~\cite{DBLP:journals/corr/abs-1903-06048} published multi-scale gradients GAN (MSG-GAN) that train multi-resolution images simultaneously. To make use of GAN for discriminative tasks, semi-supervised GAN~\cite{DBLP:journals/corr/Odena16a,NIPS2016_6125,NIPS2017_7229} jointly to train a generator and a discriminator that classify true labels for real sample and an auxiliary label for fake samples simultaneously. The scheme is good at learning a better decision boundary with only a few samples. In this work, we introduce deactivation based training with synthetic complementary samples (Deact), which is similar to semi-supervised GAN but is easier to train and implement. To make the proposed method stable, we moreover proposed decorrelation regularization to alleviate mode collapse~\cite{DBLP:journals/corr/RadfordMC15,Qi_2018_CVPR} problem in GAN.

\begin{figure*}[ht]
	\begin{center}
		%\fbox{\rule{0pt}{2in} \rule{.9\linewidth}{0pt}}
		\includegraphics[width=1.0\linewidth]{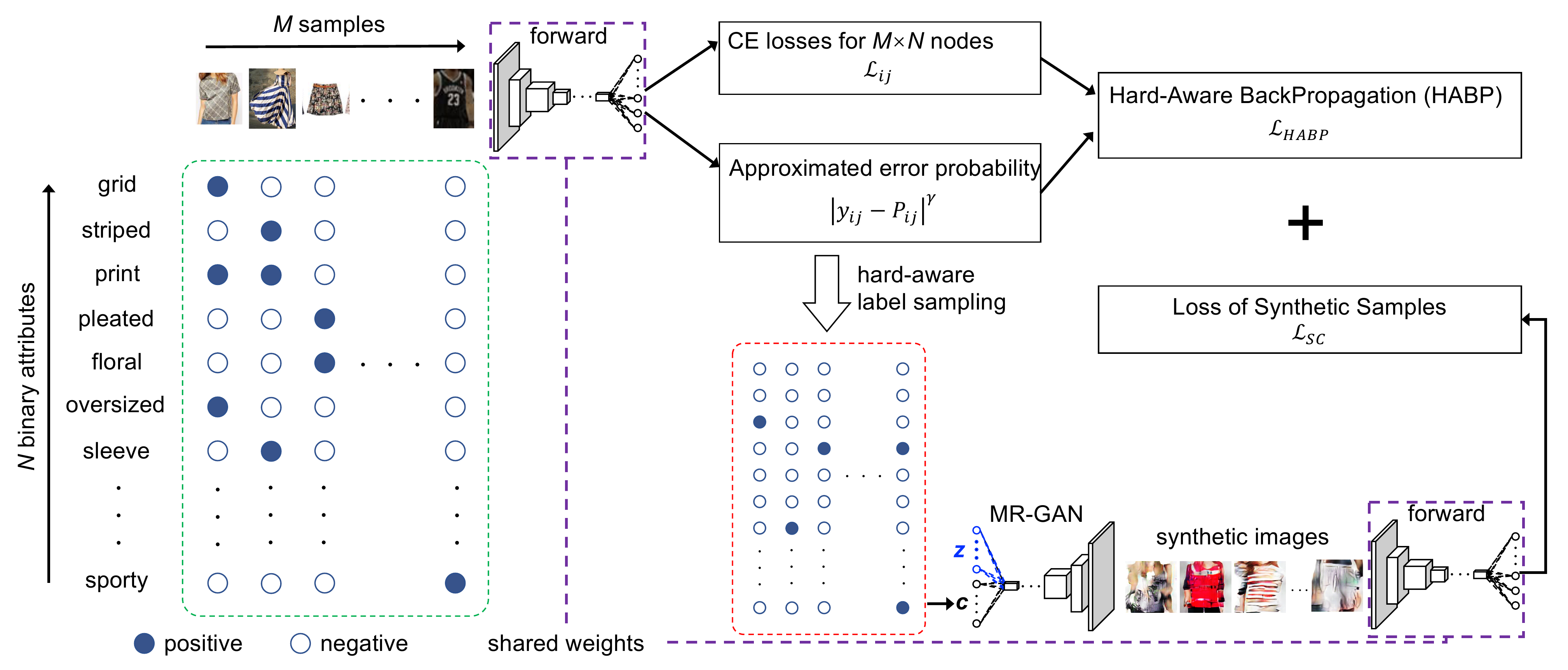}
	\end{center}
	\caption{Overall pipeline of the proposed method. HABP is calculated as the error probabilities weighted mean of cross entropy losses for all nodes. Based on the approximated error probailities, hard labels are sampled to generate synthetic complementary samples to further improve the performance. The part in purple dashed box is the network for attribute classification. }
	\label{fig:Overview}
\end{figure*}

%-------------------------------------------------------------------------
\section{Methodology}
\label{sec:methodoloy}
%-------------------------------------------------------------------------
\subsection{HABP}

The key idea of HABP is to emulate sampling losses from the output nodes. Consider a batch with $M$ samples and $N$ attributes as illustrated in Fig. \ref{fig:Overview}. After a forward pass there will be $M \times N$ output nodes, each can be calculated with labels for cross entropy (CE) loss:
\begin{equation}\label{CELoss}
\mathcal{L}_{ij}=-\log{\left(P_{ij}\right)},
\end{equation}

\noindent 
where $P_{ij}$ is the model predicted probability of target label, for the $j_{\text{th}}$ attribute of the $i_{\text{th}}$ sample in the batch. As an example, in binary classification a commonly used formula is:
\begin{equation}\label{ApproxProb}　　
P_{ij}=\begin{cases}
\sigma\left(\hat{y}_{ij}\right), & \text{if }y_{j}=1\,,  \\
1-\sigma\left(\hat{y}_{ij}\right), & \text{if }y_{j}=0\,.
\end{cases},
\end{equation}

\noindent 
where $\sigma\left(\cdot\right)$, $y$ and $\hat{y}$ are sigmoid function, ground truth label, and model output respectively. CE assumes that individual samples and attributes are equally important. When we apply CE loss on training extremely imbalanced dataset, minority attributes are always much less trained than majority attributes, resulting in much higher prediction errors.

A natural idea is to only backpropagate losses on more informative nodes. For example, a solution is to simply sample ``hard'' nodes to backpropagate losses. We borrow the idea from FL, to model the sampling probability as the probability of wrong prediction:

\begin{equation}\label{PError}　　
\left|y_{ij}-P_{ij}\right|^\gamma ,
\end{equation}

\noindent
where $\gamma$ is a tuning parameter. We then use Eq. \eqref{PError} to calculate a weighted average of losses (Eq. \eqref{CELoss}) in a batch to emulate sampling nodes for backpropagation, which we call HABP:

\begin{equation}\label{HABP}　　
\mathcal{L}_{\text{HABP}} = \frac{\sum_{i=1}^{M}\sum_{j=1}^{N}\left|y_{ij}-P_{ij}\right|^\gamma\mathcal{L}_{ij}}{\sum_{i=1}^{M}\sum_{j=1}^{N}\left|y_{ij}-P_{ij}\right|^\gamma}
\end{equation}

Note that this is equivalent to sampling nodes with the error probabilities, while it is more efficient because directly sampling suffers from the risk of missing information in unsampled nodes. Compared to FL, HABP makes hard losses more prominent and stable. Because in multi-label training, losses on hard nodes may be averaged out by the big number of attributes, particularly in the late training stage. For example, in the beginning, most attributes and samples tend to be ``hard''. As the training goes on, the ratio of ``hard'' samples will be fewer than in the beginning. If the number of attributes is large, the total losses by FL at different training stages will possibly be different in orders of magnitude, which may results in either unstable at the beginning stage or too slow learning at the late stage. More discussions with experiments will be presented in Section \ref{sec:habpvsfl}

%-------------------------------------------------------------------------
\subsection{Deactivation Training with Synthetic Complementary Samples}
\label{sec:dssgan}

As a popular re-sampling technique, semi-supervised GAN has two drawbacks: 1) Training GAN is a tricky task. There are some differences between training a GAN and training a discriminative model. For example, GAN usually requires more iterations and larger batch size~\cite{karras2018progressive,brock2018large} to achieve better image quality, which may not be optimal and necessary for training a classification model. 2) Dai \etal stated and proved that a good semi-supervised GAN requires a ``bad'' generator. Ideally, the generator should synthesize samples around but not overlapped with real samples in feature space. This is again a tricky task. 

\begin{figure}[ht]
	\begin{center}
		\includegraphics[width=0.66\linewidth]{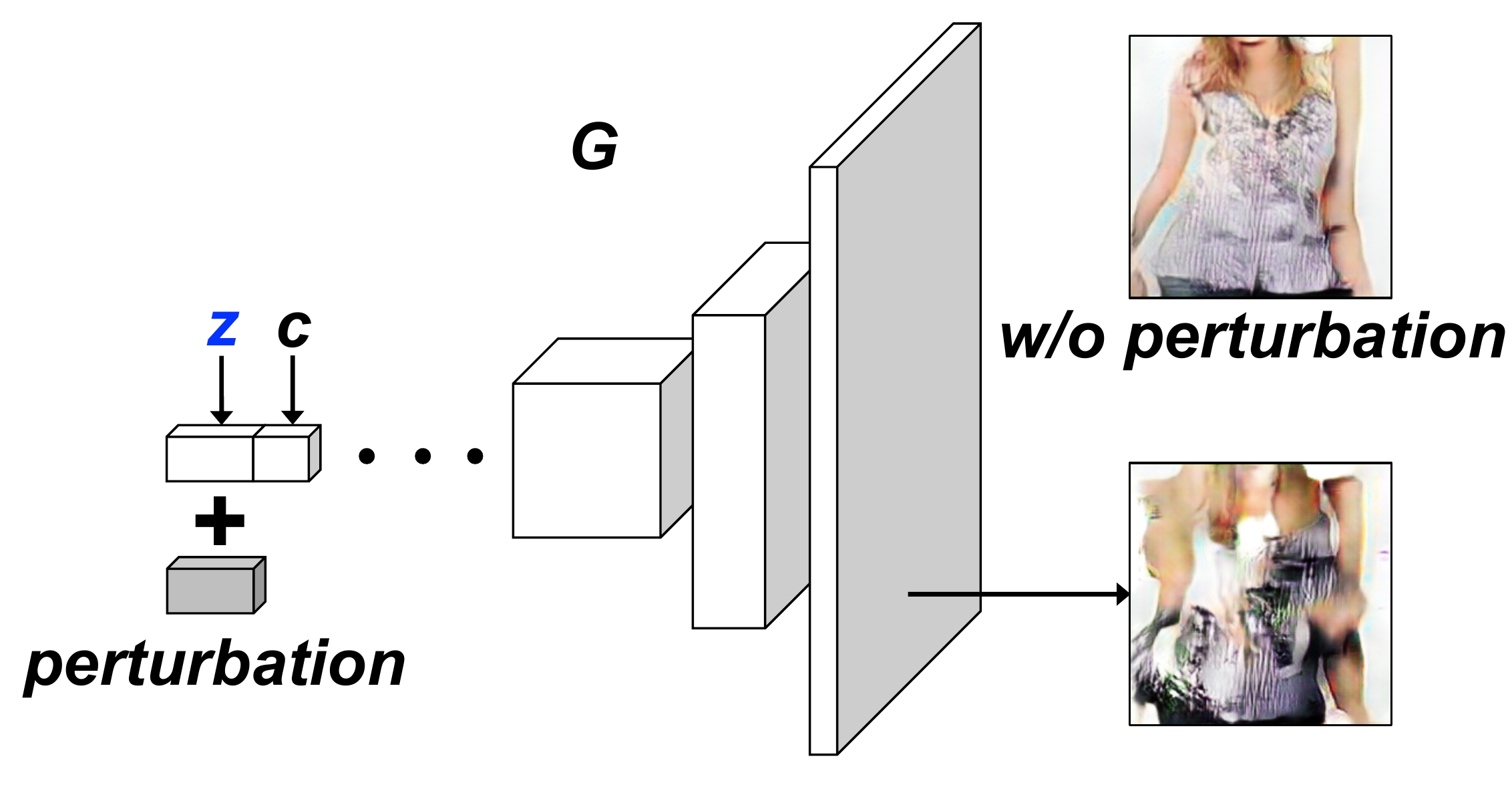}
	\end{center}
	\caption{Demonstration of generating complementary samples from a well trained GAN. $z$ represents latent noise, and $c$ represents conditional inputs. The image above is generated without perturbation.}
	\label{fig:ComplementarySample}
\end{figure}

For these reasons, we present an alternative scheme which is easier to implement and more stable in training. We first train a generator with MR-GAN with enough epochs to synthesize recognizable images. Then to make sure the generator is ``bad'' enough for semi-supervised training, we degrade the generator by adding an element-wise perturbation to the most semantic meaningful feature maps (Fig. \ref{fig:ComplementarySample}), which are the feature maps directly projected from latent space. Empirically, the perturbation should be strong enough to synthesize images that visually different from real samples as in Fig. \ref{fig:ComplementarySample}. 

To make it easier for both implementation and extendability to binary attribute case, we propose an alternative to auxiliary classifier based semi-supervised GAN. Since activating the auxiliary output for fake samples is largely equivalent to deactivating outputs for real classes, we simply pose a deactivation loss to minimize activations of real classifier outputs when training with synthetic complementary images:

\begin{equation}\label{LossSSMClass}　　
\mathcal{L}_{\text{SC},m}=\frac{1}{c}\sum_{i=1}^{c}\max(\hat{y}_i-T,0)^2,
\end{equation}

\noindent
where C is the number of classes, and $T$ is a threshold of activation. We use $T=-4.6\approx\log\left(0.01\right)$ for all the experiments that in our paper. For binary attribute classification, we want the outputs do not activate for both positive and negative, so the formula is simplified to:

\begin{equation}\label{LossSC}　　
\mathcal{L}_{\text{SC},b}=\hat{y}^2
\end{equation}

%-------------------------------------------------------------------------
\subsection{Decorrelation Regularization for MR-GAN}

\begin{figure}[ht]
	\begin{center}
		\includegraphics[width=0.8\linewidth]{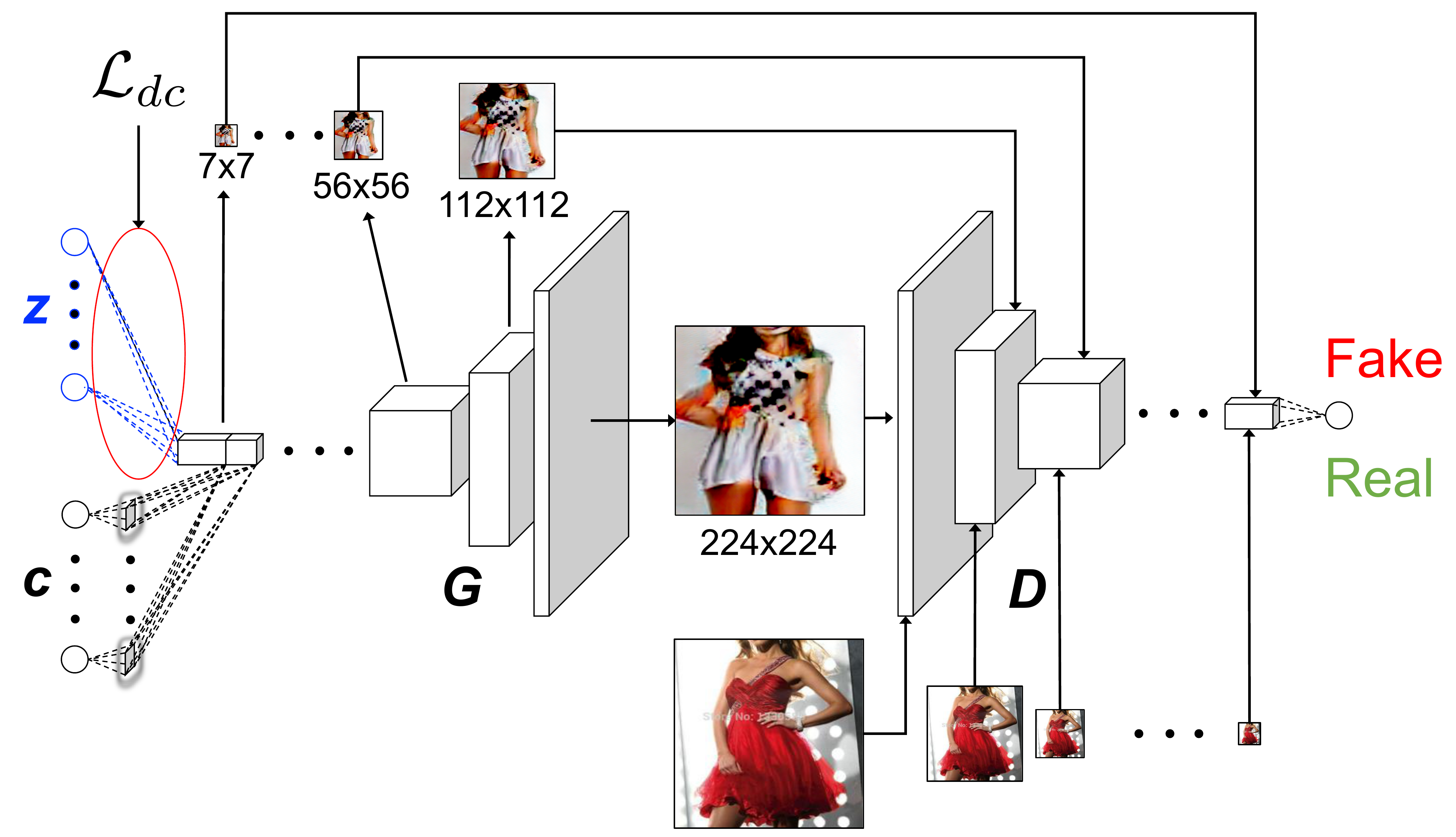}
	\end{center}
	\caption{Architecture of MR-GAN. Decorrelation regularization is applied to the weights for projecting latent noise to corresponding feature maps.}
	\label{fig:MRGAN}
\end{figure}

Aiming to synthesize high-resolution images with GAN, we employ a conditional~\cite{DBLP:journals/corr/MirzaO14} multi-resolution architecture as illustrated in Fig. \ref{fig:MRGAN}. Both generator and discriminator deal with images at different resolutions simultaneously. In Fig. \ref{fig:MRGAN} $z$ is the latent noise, and $c$ is the conditional input vector of attribute/category annotations. Each dimension corresponds to an attribute. If a positive label is sampled, the value of the corresponding dimension is set to 1. In such a structure, the higher resolution images are the refined version of lower resolution images. Thus the training is much more stable than the single resolution scheme.

As training to converge is not a problem anymore in such an architecture, we put our focus on mode collapse. Notice that the generated high-resolution images strongly depends on low-resolution images, so if we can have diversified low-resolution images, high-resolution images are not likely to fall into strong mode collapse. So we simply use a decorrelation (DC) regularization loss to decrease the correlation between latent dimensions (Fig. \ref{fig:MRGAN}). For a transposed convolution projecting $N_Z$ dim noise to $N_F$ feature maps, we denote $\mathbf{w}_{ij}$ as the filter weight of $j$th dimension of the noise to the $i$th channel of feature maps. Then we define decorrelation regularization loss as:

\begin{equation}\label{DecorrelationRegularization}　　
\mathcal{L}_{\text{DC}}=\frac{1}{N_FN_ZN_Z}\sum_{i=1}^{N_F}\sum_{j=1}^{N_Z}\sum_{k=1}^{N_Z}r\left(\mathbf{w}_{ij},\mathbf{w}_{ik}\right), 
\end{equation}

\noindent
where,

\begin{equation}\label{ZProjCorrelation}　　
r\left(\mathbf{w}_{ij},\mathbf{w}_{ik}\right)=\begin{cases}
\frac{\langle \mathbf{w}_{ij},\mathbf{w}_{ik}\rangle^2}{\langle \mathbf{w}_{ij},\mathbf{w}_{ij}\rangle\langle \mathbf{w}_{ik},\mathbf{w}_{ik}\rangle}, & \text{if }j\ne k\,,  \\
0, & \text{if }j=k\,.
\end{cases}
\end{equation}

Note that $r\left(\cdot\right)$ measures correlation as the square of cosine similarity, ranging from 0 to 1. Together with multi-resolution architecture, we call our method MR-GAN.

\subsection{Overall Training Pipeline}

With the key components above, we present our overall pipeline in Fig. \ref{fig:Overview}. The underlying idea is that semi-supervised GAN usually does not help on data with sufficient labels. So we want to train with synthetic samples only with those minority or hard attribute labels, whilst not affecting the majority or easy attributes. We implement this idea by simply sampling synthetic samples from them.

As illustrated in Fig. \ref{fig:Overview}, in each iteration we first train a batch of real samples (green dashed line box) with HABP and get the model estimated error probabilities for all labels. For each label, we update the error probability for the  $j$th attribute with an exponential moving average:

\begin{equation}\label{ErrorProbEMA}　　
S_{j,t}\left(y_{j}\right)=0.5 \overline{\left|y_{j}-P_{ij}\right|^\gamma}+0.5S_{j,t-1}\left(y_{j}\right), 
\end{equation}

\noindent
where $y_j \in \left\{0,1\right\}  $ is the label for the $j$th attribute, $S_{j,t}\left(y_j\right)$ is the being updated error of label $y_j$, $S_{j,t-1}\left(y_j\right)$ is error at last time $y_j$ showed up and $\overline{\left|y_{j}-P_{ij}\right|^\gamma}$ is the average error probability of samples that with $j$th attribute labeled as $y_j$ within a training batch. We normalize the recorded errors along each category/attribute by dividing the sum. Then they are used as the probability mass function of categorical distribution to sample hard labels. The sampled labels are used as inputs to generate the synthetic complementary samples (red dashed line box) MR-GAN. To make the deactivation based part more focused on hard labels, we again use a errors weighted average on deactivation losses for all $M\times N$ nodes:

\begin{equation}\label{LossSCAll}　　
\mathcal{L}_{\text{SC}} = \frac{\sum_{i=1}^{M}\sum_{j=1}^{N}S_j\left(y_{ij}\right)\mathcal{L}_{\text{SC},ij}}{\sum_{i=1}^{M}\sum_{j=1}^{N} S_j\left(y_{ij}\right)}
\end{equation}

The overall objective to minimize is then as follows with a tunning parameter $\lambda$ : 

\begin{equation}\label{Loss}　
\mathcal{L}=\mathcal{L}_{\text{HABP}}+\lambda\mathcal{L}_{\text{SC}}
\end{equation}

%-------------------------------------------------------------------------
\section{Experiments}

We first evaluate the proposed method on DeepFashion-C. Then more experiments on each module are further explored to verify the efficacy of the proposed method. 

\subsection{Experiments on DeepFashion-C}
\label{sec:maineval}

\begin{table*}
	\begin{center}
		\setlength{\tabcolsep}{0.72mm}
		\begin{tabular}{l|cccccccccccc|cc}
			
			\toprule[2pt]
			& \multicolumn{2}{c}{Category} & \multicolumn{2}{c}{Texture} & \multicolumn{2}{c}{Fabric} & \multicolumn{2}{c}{Shape} & \multicolumn{2}{c}{Part} & \multicolumn{2}{c}{Style}  & \multicolumn{2}{c}{All} \\
			\midrule[1pt] %line
			Top-k & Top-3 & Top-5  & Top-3 & Top-5  & Top-3 & Top-5  & Top-3 & Top-5  & Top-3 & Top-5  & Top-3 & Top-5  & Top-3 & Top-5 \\ 
			\midrule[1pt] % line
			
			FashionNet~\cite{7780493} & 82.58 & 90.17 & 37.46 & 49.52 & 39.30 & 49.84 & 39.47 & 48.59 & 44.13 & 54.02 & 66.43 & 73.16 & 45.52 & 54.61 \\
			Corbiere \etal~\cite{Corbiere_2017_ICCV} & 86.30 & 92.80 & 53.60 & 63.20 & 39.10 & 48.80 & 50.10 & 59.50 & 38.80 & 48.90 & 30.50 & 38.30 & 23.10 & 30.40 \\
			Wang \etal~\cite{Wang_2018_CVPR} & \textbf{90.99} & \textbf{95.78} & 50.31 & 65.48 & 40.31 & 48.23 & 53.32 & 61.05 & 40.65 & 56.32 & \textbf{68.70} & \textbf{74.25} & 51.53 & 60.95 \\
			OHEM~\cite{Shrivastava_2016_CVPR} & 89.66 & 95.28 & 58.19 & 67.60 & 45.20 & 55.61 & 57.83 & 67.01 & 45.09 & 55.21 & 33.33 & 41.79 & 48.40 & 58.02 \\
			FL~\cite{Lin_2017_ICCV} & 90.38 & 95.51 & 59.63 & 69.15 & 47.95 & 58.61 & 61.26 & 70.16 & 50.23 & 60.16 & 36.22 & 44.76 & 51.46 & 61.10 \\
			Weighted FL~\cite{Sarafianos_2018_ECCV} & 90.32 & 95.39 & 58.52 & 68.26 & 47.65 & 58.07 & 60.77 & 69.62 & 50.37 & 60.74 & 36.79 & 45.60 & 51.31 & 61.01 \\
			
			\midrule[1pt] % line
			Weighted CE-A & 90.20 & 95.25 & 58.74 & 68.96 & 47.31 & 57.65 & 60.35 & 69.54 & 48.59 & 59.37 & 37.20 & 45.68 & 50.79 & 60.68 \\
			Weighted CE-B & 88.21 & 93.72 & 58.03 & 67.98 & 45.30 & 56.27 & 58.82 & 68.68 & 47.23 & 58.53 & 33.21 & 42.53 & 48.95 & 59.30 \\
			\midrule[1pt] % line
			Baseline & 89.93 & 95.20 & 57.08 & 66.72 & 43.96 & 54.19 & 56.79 & 65.98 & 44.36 & 54.09 & 33.10 & 41.40 & 47.46 & 57.00 \\
			Deact only & 90.93 & 95.73 & 58.52 & 68.26 & 46.38 & 56.82 & 59.09 & 68.08 & 47.66 & 57.84 & 35.66 & 44.28 & 49.84 & 59.54 \\
			HABP only & 89.96 & 94.89 & 60.34 & 70.06 & 48.73 & 59.65 & 61.44 & 70.70 & 50.73 & 61.09 & 37.69 & 46.23 & 52.17 & 62.07 \\
			FL+Deact  & 89.92 & 95.00 & 60.38 & 70.23 & 49.01 & 59.53 & 61.52 & 70.32 & 51.07 & 61.14 & 37.85 & 46.57 & 52.36 & 62.06 \\
			HABP+Deact & 90.06 & 95.04 & \textbf{60.87} & \textbf{70.54} & \textbf{49.40} & \textbf{59.88} & \textbf{61.97} & \textbf{70.80} & \textbf{51.39} & \textbf{61.82} & 38.61 & 46.99 & \textbf{52.82} & \textbf{62.49} \\
			\bottomrule[2pt] % line
		\end{tabular}
	\end{center}
	\caption{Results of category and attribute classification (\%)}
	\label{table:Evaluation-DFC}
\end{table*}

\begin{table}
	\begin{center}
		\setlength{\tabcolsep}{1.mm}
		\begin{tabular}{c|c|c|c|c|c|c}
			\toprule[2pt]
			& Texture & Fabric & Shape & Part & Style & All \\
			\midrule[1pt] % line
			CRL~\cite{8353718} & 55.37 & 55.02 & 55.22 & 53.90 & 53.75 & 54.56 \\
			W-CE-B & 76.88 & 77.10 & 81.86 & 76.81 & 72.23 & 76.73 \\
			HABP & 77.10 & 77.58 & 82.26 & 77.31 & 73.26 & 77.29 \\
			Deact & 78.59 & 78.31 & \textbf{83.83} & 78.57 & 74.20 & 78.45 \\
			Ours & \textbf{78.69} & \textbf{78.82} & 83.80 & \textbf{79.23} & \textbf{74.26} & \textbf{78.74} \\
			\bottomrule[2pt] % line
		\end{tabular}
	\end{center}
	\caption{Class-balanced accuracy on DeepFashion-C (\%)}
	\label{table:EvalCBA}
\end{table}

\textbf{Dataset}. The 1000 attributes of DeepFashion-C~\cite{7780493} are divided into 5 groups by the authors, characterizing texture, fabric, shape, part, and style. We follow the official split by DeepFashion-C, more specifically, 209,222 training samples, 40,000 validation samples, and 40,000 test samples. The validation set is only used to make sure there was no overfitting.

\textbf{Evaluation Metrics}. Two evaluation metrics and the corresponding settings are used: 1) top-k recall/accuracy. For binary attribute prediction, we calculate top-k recall following~\cite{DBLP:journals/corr/GongJLTI13}, which is obtained by ranking the classification scores and determine how many attributes have been matched in the top-k list for each group. For category classification, top-k classification accuracy is calculated; 2) To further prove the effectiveness and flexibility of our approach, we also conduct experiments evaluating the class-balanced accuracy for attributes, which is calculated by averaging accuracy for both positive and negative labels attribute-wise~\cite{8353718}.

\textbf{Comparsions}. 
For attribute and category classification, we compared our method with recently published results~\cite{7780493,Corbiere_2017_ICCV,Wang_2018_CVPR}, and our re-produced results of popular hard-aware methods including OHEM~\cite{Shrivastava_2016_CVPR} focal loss (FL)~\cite{Lin_2017_ICCV}, and weighted FL for multi-label dataset~\cite{Sarafianos_2018_ECCV}. Weighted FL weight loss of each attribute with $w_c=e^{-a}$, where a is the prior attribute distribution. For OHEM we tried different ratio (0.5, 0.33, 0.17, 0.1) of hard nodes, and select the best result with the ratio of 0.17. For FL based methods, we use the same $\gamma$=1.2 as we used for HABP. As we discussed in Section \ref{sec:methodoloy}, without an output dependent normalization term, FL may result in either unstable at the beginning stage or too slow learning in the late stage. To avoid a low performance of FL by either case and make the comparison more sensible, we tried different base learning rate for FL. In our experiments, we found that for top-k recall/accuracy $lr=0.2$ gives the best result using FL. Similarly, we run experiments for weighted FL and report the best results with $lr=0.15$. We also tried a commonly used strategy to weight the positive/negative ratio for binary cross entropy loss:

\begin{equation}\label{WBCELoss}　　
w_jy_j\log\left(\sigma\left(\hat{y}_j\right)\right)
+\left(1-y_j\right)\log\left(\sigma\left(1-\hat{y}_j\right)\right),
\end{equation}

\noindent
where $w_n$ is a weight depends on positive/negative ratio of a given attribute. Denoting $n_{\text{pos},j}$ and $n_{\text{neg},j}$ as number of positive and negative labels for the $j_{\text{th}}$ attribute among $N$ attributes, we tried two ways for the weight\footnote{We didn't use the negative to positive ratio as $w_j$ to balance the CE loss. Because we have tried different settings with this, and observed either numerical instability or very bad performance due to the very large variation of the ratio (6.7$\sim$7471.2). }: \textbf{A: weight each attribute adaptively}. $w_j=\log\left(n_{\text{neg},j}/n_{\text{pos},j}\right)$; \textbf{B: one weight for all attributes}. $w_j=\sum_{j=1}^{N} n_{\text{neg},j}/\sum_{j=1}^{N}n_{\text{pos},j}$. 

\textbf{Implementation Details}. 1) For top-k recall/accuracy, the base model we used is an imagenet pre-trained VGG-16~\cite{DBLP:journals/corr/SimonyanZ14a}. We replace the fully-connected layers by two $3\times 3$ convolutions without padding. The first convolution outputs 2048 channels, and the second outputs 4096 channels. Each convolution is followed by a ReLU activation. Then the 4096 channels are reduced to a vector by average pooling. A dropout with the probability of 0.5 is followed to avoid over-fitting. Final output for category and binary attributes are fully-connected layers with 50 and 1000 outputs respectively. We view attribute classification as 1000 binary classification tasks, and category classification as a multi-class task, such that the loss weight we used for category classification is the same as every single task in attribute classification. We train the network 15 epochs with mini-batch of 16 images in all experiments. Each image is cropped with the ground truth bounding box and resized to $224\times 224$. For the first 6 epochs, the learning rate is 0.01, then it is decreased by a factor of 10 every 3 epochs. $\gamma$ for HABP is set to 1.2 for all experiments. Loss from synthetic complementary samples is added with a weight of 1e-4, and $\sigma_p$ for semantic feature perturbation is set to 1.5. For training efficiency, deactivation loss is computed every 20 iterations. 2) For class-balanced accuracy, we follow ~\cite{8353718} by using ResNet50~\cite{7780459} as the network. We found that the best result is achieved by using weighted CE-B described in the last paragraph as the base loss term $\mathcal{L}_{ij}$ in eq. \ref{HABP}. For experiments, the weight for deactivation loss is 0.001, and $\gamma$=0.1. Other settings remain the same as the experiments for top-k recall/accuracy. 

\textbf{Results}. As mentioned before, error probability strongly depends on the number of positive labels. We first verified this and the effectiveness of our approach on reducing prediction errors of minority data. For the convenience of visualization, we compute the average of predicted positive probability $\sigma(\hat{y})$ for all positive labels instead of error probability. Comparisons between two well-trained models with CE loss and our method on both train and test set are illustrated in Fig. \ref{fig:ppcorr}. From the figure, we can see that our method significantly reduced errors of positive labels, particularly for minority labels.

The evaluation results using top-k recall/accuracy on test set are summarized in Table \ref{table:Evaluation-DFC}. From the table, our overall performance on attribute classification out-performs all others including the current state-of-the-art~\cite{Wang_2018_CVPR}. The category classification result is also better than most of the others. We also observed that with only HABP, the result surpass FL, weighted FL, and OHEM, which are methods with similar spirits. This proved the better stability of HABP. To understand this, consider that if the number of hard nodes is only a few in a batch, both FL and OHEM will result in a small loss that does not contribute much to gradients, while HABP constantly backpropagates a stable total loss from hard nodes no matter how many hard nodes in a batch. Together with HABP, deactivation based training with synthetic complementary samples further improves the final result, as demonstrated in the lower part of Table \ref{table:Evaluation-DFC}. Note that our method only used attribute annotations, while in both \cite{7780493} and \cite{Wang_2018_CVPR} landmark annotations are used to enhance the attribute classification. 

\textbf{An ablation study} is also presented in the lower part of Table \ref{table:Evaluation-DFC}. By independently activate HABP and synthetic complementary samples, we found the two techniques both improves over baseline. We also tried to replace HABP with FL in the pipeline. It achieves a better result than both baseline and deactivation based training with synthetic complementary samples. Yet it is still lower than our proposed pipeline, which further proves the advantage of HABP over FL.

\textbf{The experimental results with class-balanced accuracy} in Table \ref{table:EvalCBA} further shows the flexibility and superiority of the proposed method. We observed that both HABP and deactivation loss improve the performance by some margin. Unlike the baseline with CE loss, by using the settings of weighted CE-B, training with synthetic complementary samples contributes more than HABP. We think a future work worth to study is how to optimally combine HABP and Deact given a specific task.

\subsection{HABP vs. FL}
\label{sec:habpvsfl}

\begin{figure}[t]
	\begin{center}
		\includegraphics[width=0.66\linewidth]{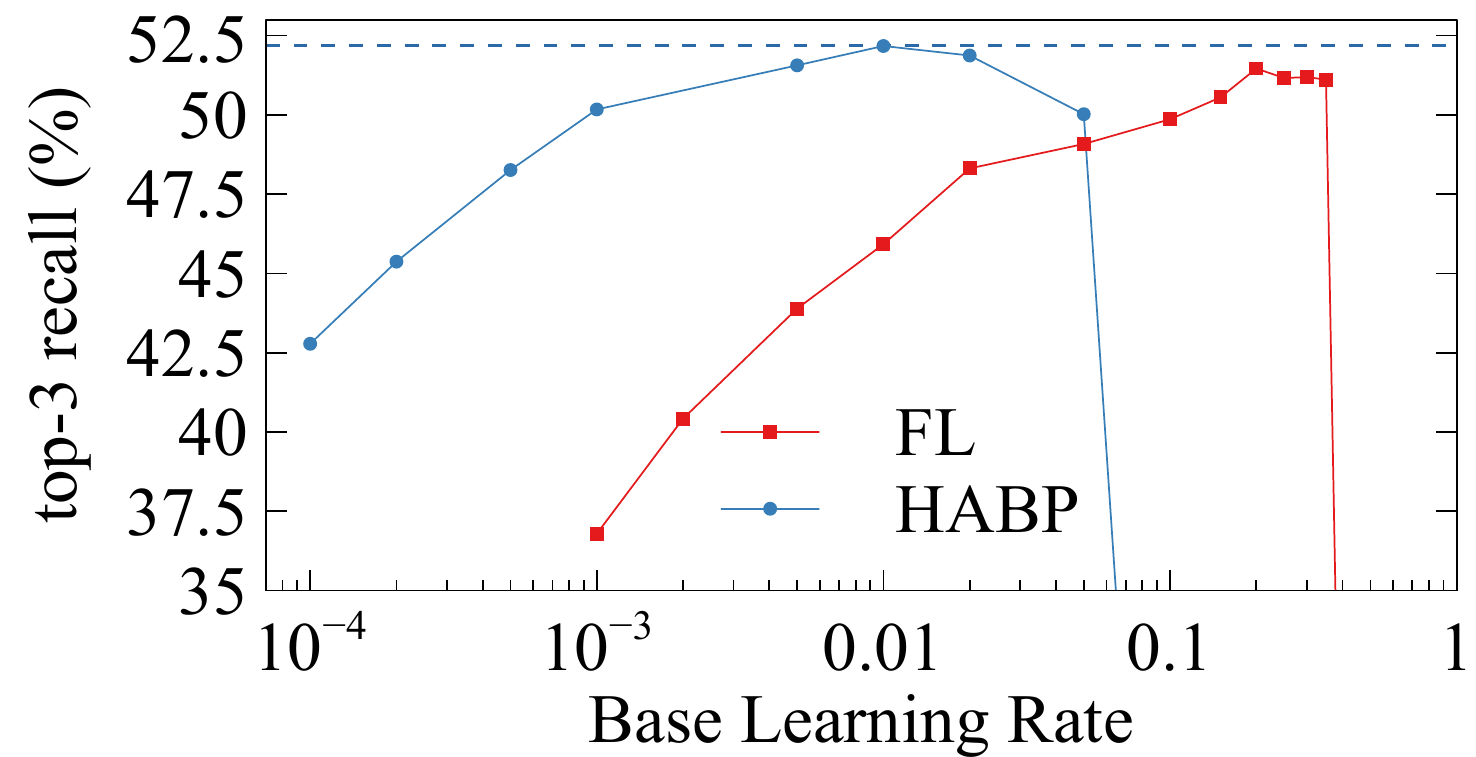}
	\end{center}
	\caption{Top-3 recall of HABP \& FL under different learning rate. HABP demonstrates better stability and performance than FL. }
	\label{fig:habpvsfl}
\end{figure}

\begin{figure}[t]
	\begin{center}
		\includegraphics[width=0.8\linewidth]{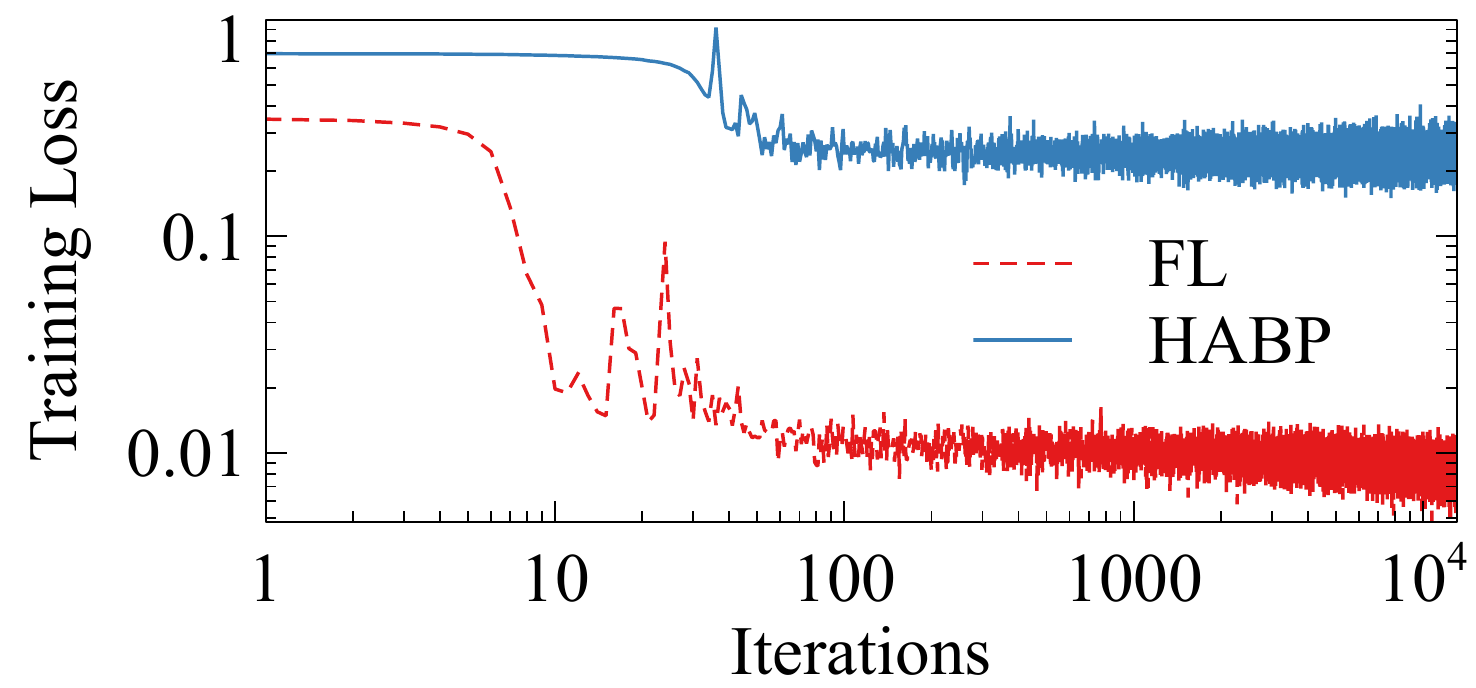}
	\end{center}
	\caption{ Training loss of attributes in the $1_{st}$ epoch. HABP  }
	\label{fig:habpfltraincurve}
\end{figure}

In Table \ref{table:Evaluation-DFC} we already verified the better performance of our method over other popular choices. In this section, we focus on the comparison between HABP and FL by more experiments. As we already mentioned in Section \ref{sec:maineval}, we tune the base learning rate for FL to avoid either too low learning or numerical instability. With the sample experimental settings for top-k recall/accuracy in Table. \ref{table:Evaluation-DFC}, we demonstrate the top-3 recall for attributes by both HABP and FL under different base learning rate in Fig. \ref{fig:habpvsfl}. Compared to FL, HABP demonstrates not only better performance (as shown in the blue dashed line), but also less sensitive to base learning rate. 

We also plotted the training loss for attributes in the first epoch with the experiment setting for Table. \ref{table:Evaluation-DFC}. From Fig. \ref{fig:habpfltraincurve} we can see that loss calculated by HABP keeps prominent as training goes, while the loss by FL at the beginning is almost two orders of magnitude larger than the loss at the end of the first epoch. This sensitive behavior of FL limited its performance because a too large learning rate may result in convergency issues, while a smaller learning rate may not be able to learn parameters in the late stage. 

\begin{figure}[t]
	\begin{center}
		\includegraphics[width=0.8\linewidth]{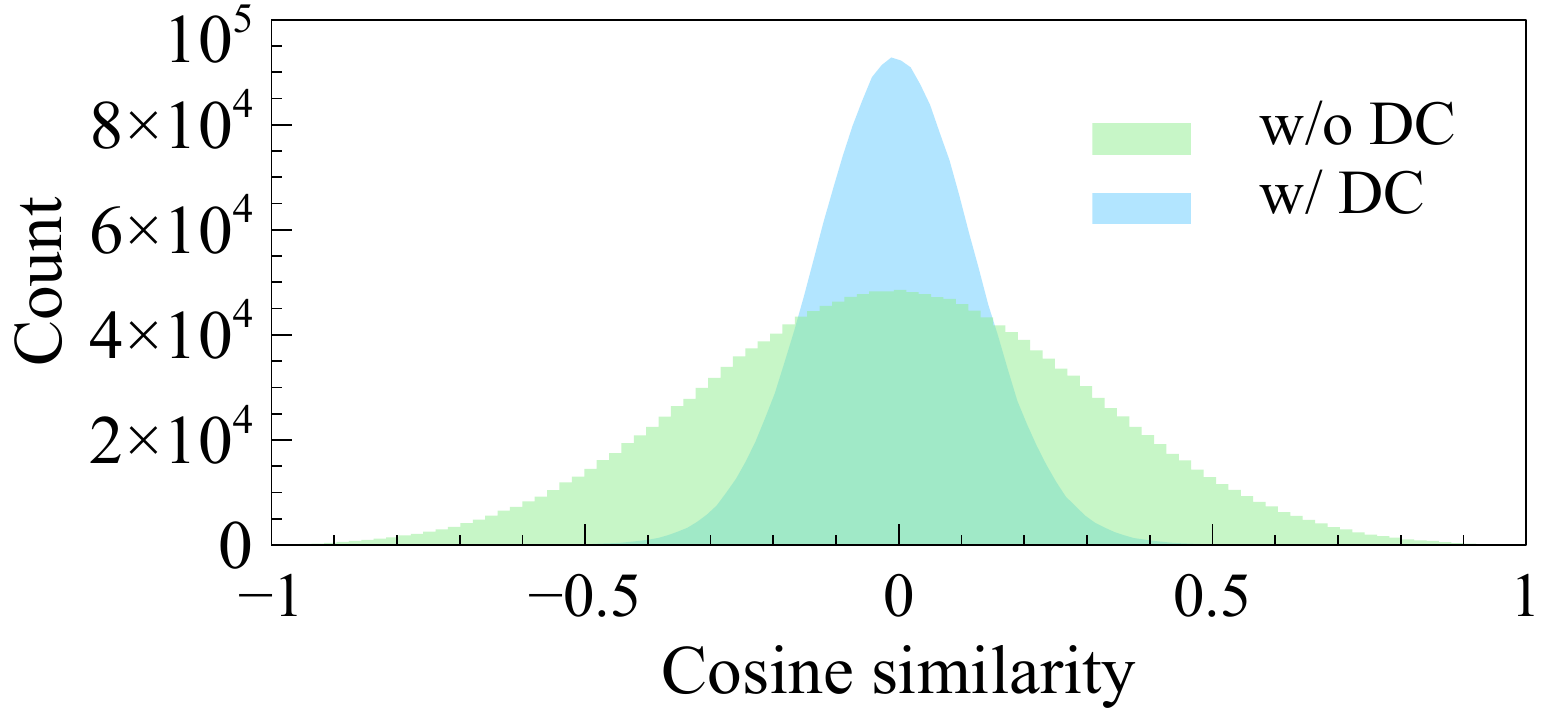}
	\end{center}
	\caption{Correlations between weights w/ and w/o decorrelation regularization}
	\label{fig:CorrDist}
\end{figure}

\begin{figure}[ht]
	\begin{center}
		\includegraphics[width=0.8\linewidth]{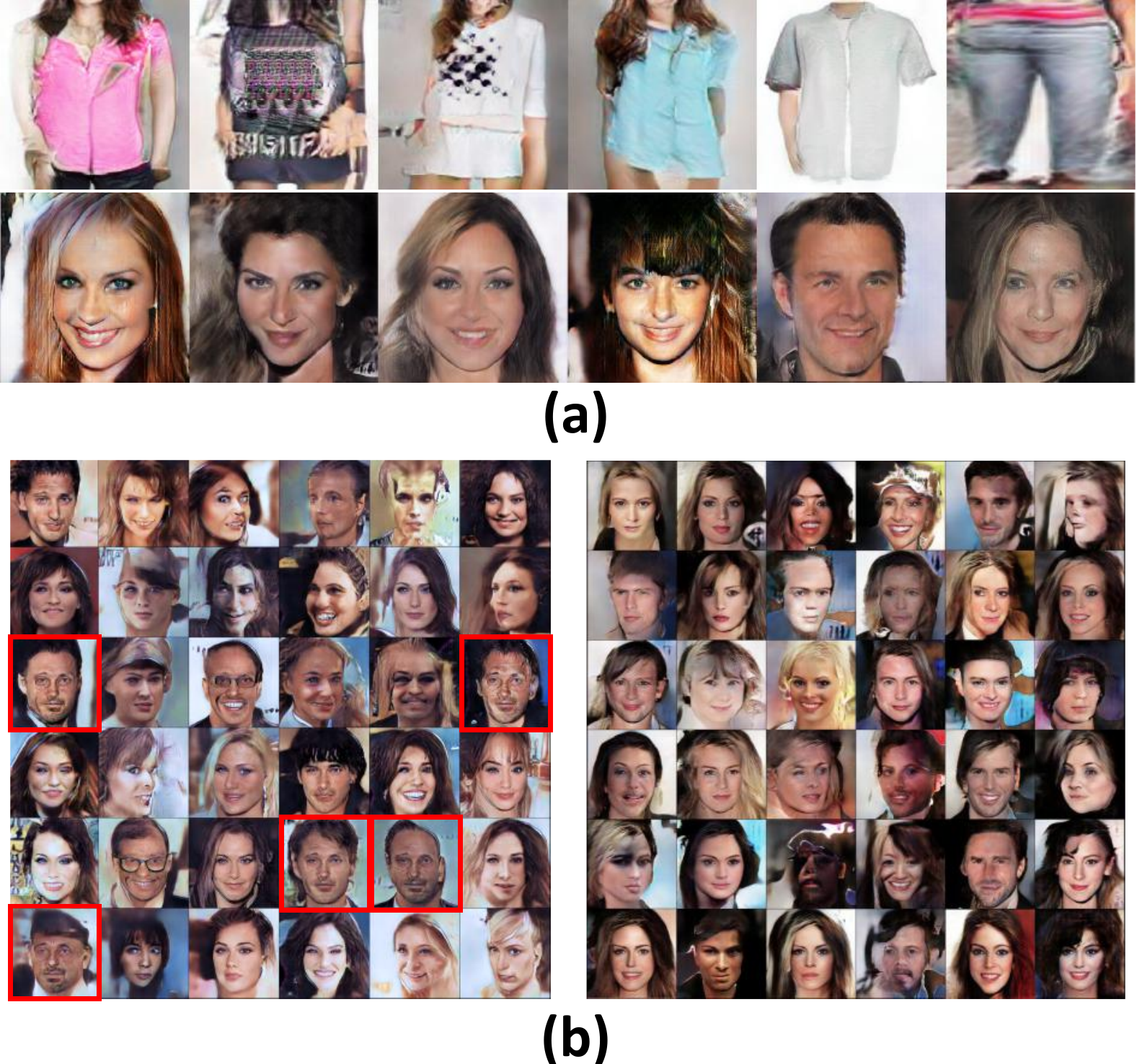}
	\end{center}
	\caption{Images generated using MR-GAN. (a) \textbf{Upper}: $224 \times 224$ samples generated using DeepFashion-C. \textbf{Lower}: $512 \times 512$ samples generated using CelebA-HQ. (b) Random samples w/ (\textbf{right}) and w/o (\textbf{left}) decorrelation regularization loss.}
	\label{fig:SynthImages}
\end{figure}

\subsection{More Experiments of Deact}

We further independently verified the validity of the proposed deactivation based training on MNIST classification in this section. The network we used is a LeNet-5~\cite{726791} with ReLU activations. We train a subset of randomly sampled training data with different sizes ranging from 25$\sim$1000. The total number of training samples is set to 500k, with a batch size of 64. SGD optimizer is used with a learning rate of 0.01 and momentum of 0.9. The weight of deactivation loss is $\lambda$=0.05 for all experiments. To generate complementary samples we use MR-GAN with the following sequential operations as the building block: {\it (transposed) convolution}$\rightarrow${\it batch normalization}$\rightarrow${\it leaky ReLU}. The number of channels inversely depends on the size of feature maps, and for generating highest resolution images we set it to 64. MR-GAN is also trained with the same subset. In total 1M samples are trained with a batch size of 128. The $\sigma$ of Gaussian noise used to perturb feature maps, and the improvements with our proposed method are summarized in Table \ref{table:DSSGAN-CErrors}.

\begin{table}
	\begin{center}
		\begin{tabular}{c|c|c|c}
			\toprule[2pt]
			\# of samples & LeNet & Deactivation & AC \\
			\midrule[1pt]
			25 & 44.12\% & \textbf{40.06\%}($\sigma_p$=3.6) & 43.71\% \\
			50 & 29.73\% & \textbf{26.22\%}($\sigma_p$=2.2) & 27.65\% \\
			100 & 14.53\% & \textbf{12.14\%}($\sigma_p$=2.4) & 14.33\% \\
			500 & 4.88\% & \textbf{4.58\%}($\sigma_p$=2.4) & 5.01\% \\
			1000 & 3.93\% & \textbf{3.72\%}($\sigma_p$=2.2) & 4.01\% \\
			\bottomrule[2pt]
		\end{tabular}
	\end{center}
	\caption{Classification errors on MNIST~\cite{726791}}
	\label{table:DSSGAN-CErrors}
\end{table}

\begin{table}
	\begin{center}
		\begin{tabular}{c|c|c}
			\toprule[2pt]
			& DeepFashion-C~\cite{7780493} & CelebA-HQ~\cite{karras2018progressive}\\
			\midrule[1pt]
			w/o DC & 28.51 & 26.93 \\
			w/ DC & \textbf{27.28} & \textbf{22.16} \\
			\bottomrule[2pt]
		\end{tabular}
	\end{center}
	\caption{FID with and without decorrelation regularization on DeepFashion-C and CelebA-HQ. The lower the better.}	\label{table:MRGAN-arch}
\end{table}

\subsection{Effectiveness of Decorrelation Regularization and MR-GAN}

We validate the effectiveness of MR-GAN with DeepFashion-C and CelebA-HQ~\cite{karras2018progressive}. In training, images at different resolutions are generated by one forward pass, whilst multiple forwards with discriminator are needed for generating corresponding outputs. Due to the strong stability of MR-GAN, we simply use vanilla GAN loss~\cite{NIPS2014_5423} for training. The proposed decorrelation regularization is added to the loss of G with a weight of 2e-6 for all experiments. For discriminator, we calculate the mean over losses for multiple resolutions as the final loss. 
\textbf{Implementation Details}. We crop each image with ground truth bounding box provided by DeepFashion-C, and resize to $224\times 224$. Adam~\cite{DBLP:journals/corr/KingmaB14} optimizer is used for both G and D with a learning rate at 1e-4. We use 32 as the number of channels for generating the highest resolution images, and the maximum number of channels is set to 512. The network is trained for 30 epochs with the batch size of 128, on train set only. To further validate the MR-GAN's ability to synthesizes higher resolution images, we also experimented with CelebA-HQ dataset, which contains 30k face images at $1024 \times 1024$. We build the network for images from $4 \times 4$ to $512\times 512$. The number of channels for the largest image is set to 12, and the number of channels is limited up to 384. For CelebA-HQ we train without conditional inputs for 50 epochs, with mini-batch of 64 images.

We computed Fréchet Inception Distance~\cite{NIPS2017_7240} (FID) from 30k images on the last 10 epochs for both datasets, and pick the smallest FID. For DeepFashion-C, the labels are sampled with prior distribution from the train set. The results are summarized in table \ref{table:MRGAN-arch}, showing that the image samples using both datasets with decorrelation regularizations are better than without it.

The cosine similarities between the transposed convolution kernels that projecting latent noises are calculated and demonstrated in Fig. \ref{fig:CorrDist}. The correlations between weights are clearly reduced by decorrelation regularization loss as shown. Sample images generated with MR-GAN are illustrated in Fig. \ref{fig:SynthImages}(a). In Fig. \ref{fig:SynthImages}(b), left is random samples without decorrelation loss, we can see very similar faces labeled with red boxes, while this is not observed in the right image with decorrelation regularization loss.

%------------------------------------------------------------------------
\section{Conclusion}

We have proposed a pipeline to make use of ``hard'' data with two techniques from the view of cost-sensitive learning and the view of re-sampling respectively. It consists of HABP that effectively and adaptively learning with hard data, and deactivation based training with synthetic complementary samples that is more stable to train and easier to implement. HABP focus on positive minority data, whilst deactivation based training helps to learn a better decision boundary by deactivating complementary samples for minority data. The two components can either be combined or separately used depending on the specific metric. Along with the pipeline, we also presented a decorrelation regularization loss for training a multi-resolution GAN. Evaluations are performed on a large scale fashion dataset and related datasets. Overall our method achieves the state-of-the-art for attribute classification. At the same time, from the observations in experiments, we believe how to optimally combine the components we proposed will be a topic that worth future studies.

\bibliographystyle{unsrt}  
\bibliography{references}  %%% Remove comment to use the external .bib file (using bibtex).

\end{document}